\documentclass[review]{elsarticle}

\usepackage{lineno,hyperref}
\usepackage{mathrsfs}

\usepackage{algpseudocode}  
\usepackage{amsmath}  
\usepackage{amsfonts}

\usepackage{pgfplots}
\usepackage{caption}
\usepackage{subcaption}
\usepackage{microtype}
\usepackage{multirow}
\usepackage{colortbl}
\usepackage{amssymb}
\usepackage[linesnumbered,ruled,boxed,norelsize,vlined,commentsnumbered]{algorithm2e}
\usepackage{pifont}

\usepackage{amsmath}
\usepackage{amsthm}
\usepackage{adjustbox}
\usepackage{booktabs}

\newcommand{\pgen}{p_{\text{gen}}}
\newcommand{\refine}{p_{\text{refine}}}

\newcommand{\tbl}[1]{\textcolor{black}{#1}}

\journal{}

\begin{document}

\begin{frontmatter}

\title{Contrastive Prompt Clustering for Weakly Supervised Semantic Segmentation}
		
		\author[firstaddress,secondaddress]{Wangyu Wu}\ead{wangyu.wu@liverpool.ac.uk}
            \author[thirdaddress]{Zhenhong Chen}\ead{zcheh@microsoft.com}
            \author[Faddress]{Xiaowen Ma}\ead{xwma@zju.edu.cn}
            \author[Faddress]{Wenqiao Zhang}\ead{wenqiaozhang@zju.edu.cn}
            \author[firstaddress,secondaddress]{Xianglin Qiu}\ead{Xianglin.Qiu20@student.xjtlu.edu.cn}
            \author[firstaddress,secondaddress]{Siqi Song}\ead{Siqi.Song22@student.xjtlu.edu.cn}

            \author[secondaddress]{Xiaowei Huang}\ead{xiaowei.huang@liverpool.ac.uk}
		\author[firstaddress]{Fei Ma\corref{mycorrespondingauthor}}\ead{fei.ma@xjtlu.edu.cn}

            \author[firstaddress]{Jimin Xiao\corref{mycorrespondingauthor}}
		\cortext[mycorrespondingauthor]{Corresponding authors} \ead{jimin.xiao@xjtlu.edu.cn}
		
		\address[firstaddress]{Xi'an Jiaotong-Liverpool University, Suzhou, China}
		\address[secondaddress]{University of Liverpool, Liverpool, UK}
            \address[thirdaddress]{Microsoft, Redmond, USA}
            \address[Faddress]{Zhejiang University, Hangzhou, China}

\begin{abstract}
Weakly Supervised Semantic Segmentation (WSSS) with image-level labels has gained attention for its cost-effectiveness. Most existing methods emphasize inter-class separation, often neglecting the shared semantics among related categories and lacking fine-grained discrimination. To address this, we propose Contrastive Prompt Clustering (CPC), a novel WSSS framework. CPC exploits Large Language Models (LLMs) to derive category clusters that encode intrinsic inter-class relationships, and further introduces a class-aware patch-level contrastive loss to enforce intra-class consistency and inter-class separation. This hierarchical design leverages clusters as coarse-grained semantic priors while preserving fine-grained boundaries, thereby reducing confusion among visually similar categories. Experiments on PASCAL VOC 2012 and MS COCO 2014 demonstrate that CPC surpasses existing state-of-the-art methods in WSSS.

\end{abstract}


\end{frontmatter}

%\linenumbers
\section{Introduction} \label{sec:intro}

Weakly supervised semantic segmentation (WSSS) aims to produce pixel-level semantic predictions using only image-level annotations~\cite{wu2025adaptive,wu2024image,wanwu3,wanwu4}. This paradigm greatly reduces the cost and labor associated with fine-grained labeling, making it an appealing alternative to fully supervised methods. Despite the weak supervision, WSSS methods strive to generate high-resolution segmentation masks where each pixel is assigned an appropriate class label.

\begin{figure}[t]
\centering
%\includesvg[width=0.8\linewidth]{figure/idea.svg}
\includegraphics[width=1.0 \linewidth]{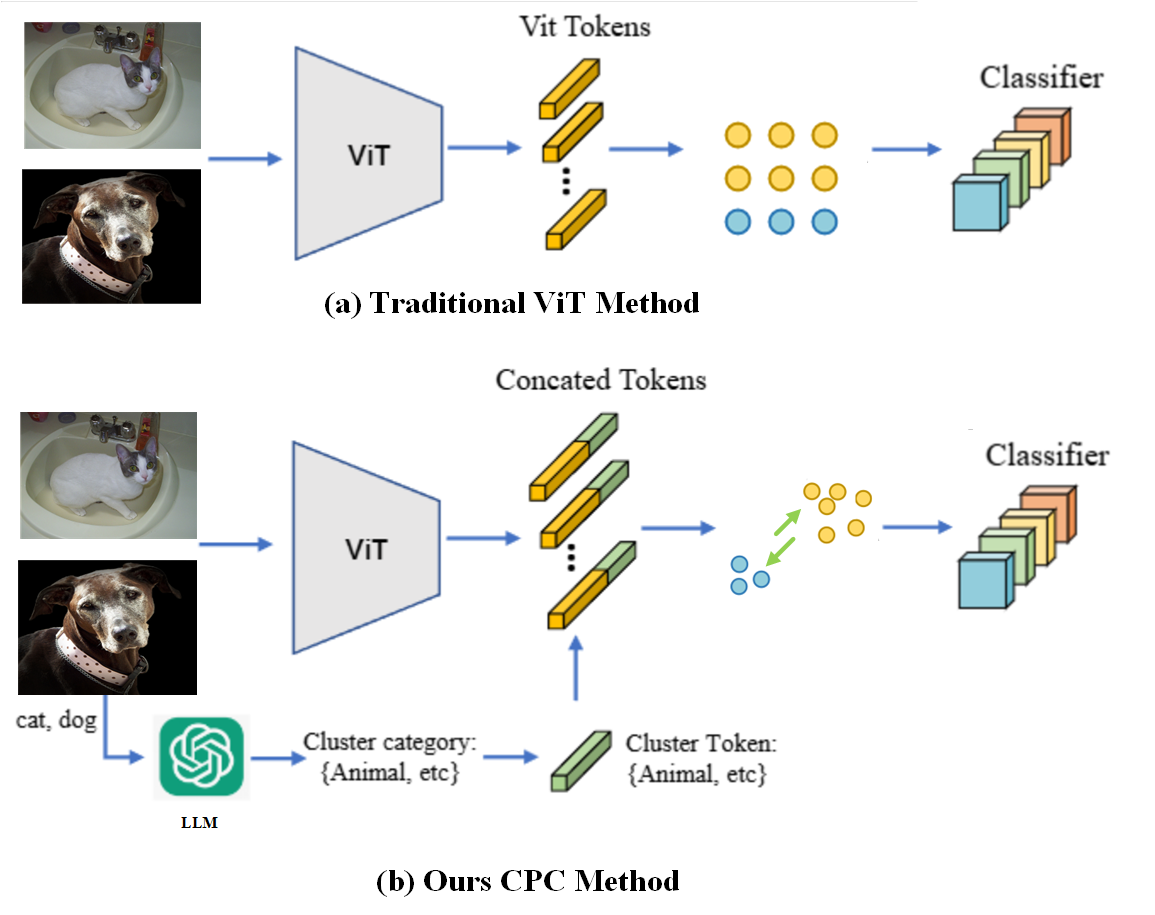}

\caption{(a) In traditional ViT-based WSSS methods, only the ViT tokens representing image patches are used for classification. 
(b) In our CPC, we leverage a Large Language Model (LLM) to determine the cluster category of each image and generate a cluster token that encodes shared class information. 
This cluster token, combined with contrastive learning, enhances the ViT patch tokens.}
\label{fig:idea}

\end{figure}
Recently, many WSSS methods~\cite{kweon2024sam,yin2023semi,chen2024region,li2023high-resolution,wanwu2} adopt the Vision Transformer (ViT) architecture~\cite{dosovitskiy2020image,wanwu1,wu2024top}, due to its capability to capture global contextual dependencies through self-attention mechanisms. These mechanisms allow ViT to model long-range interactions across an image, thereby improving semantic coherence. However, the self-attention in ViT also tends to act as a low-pass filter, diminishing the variance of local features and leading to excessive smoothing~\cite{yin2024classm}. This effect often suppresses high-frequency details that are critical for preserving fine-grained boundaries and intra-object distinctions, which ultimately weakens the discriminability of patch-level features and harms segmentation quality. In addition, existing WSSS approaches~\cite{li2024high-fidelity,yin2024classm,guo2024dual-hybrid} typically emphasize maximizing inter-class separation to reduce class confusion. While this strategy can enhance overall discriminative power, it often overlooks the valuable shared semantic information among visually or semantically similar categories. We argue that mining and leveraging such shared information is essential for capturing the latent structure of the semantic space. To this end, we propose to group related classes into semantic clusters, and inject their shared features into patch representations to mitigate the over-smoothing effect and reduce ambiguity, as illustrated in Fig.~\ref{fig:idea}.

To effectively identify and exploit inter-category relationships, we propose \textbf{Contrastive Prompt Clustering (CPC)}. Our method leverages Large Language Models (LLMs) to generate prompts that encode semantic affinities between categories, forming prompt-guided clusters that serve as high-level relational priors to guide recognition of shared semantics across similar classes. \tbl{To further enhance class-level discriminability, we introduce a class-aware patch-level contrastive module that enforces intra-class compactness and inter-class separability, pulling embeddings of the same class closer while pushing apart those of different classes. While prompt clustering captures coarse-grained semantic similarities, the contrastive objective imposes fine-grained visual constraints, even within semantically related clusters. This hierarchical design allows the model to jointly benefit from shared semantic priors and class-specific distinctions.}

\textbf{Our main contributions are fourfold}:
\begin{itemize}
    \item We propose a ViT-based WSSS framework that integrates prompt-derived category clusters and class labels into the segmentation process, enriching the model's semantic representation.
    \item We develop an automated semantic clustering method based on GPT-generated prompts, enabling the model to discover and utilize inter-category relationships without external supervision.
    \item \tbl{We introduce a class-aware patch-level contrastive learning objective that enhances fine-grained class discrimination while remaining compatible with the prompt-guided clustering strategy.}
    \item We demonstrate the effectiveness of our approach on both the PASCAL VOC 2012 and MS COCO 2014 datasets, where our method achieves superior performance compared to state-of-the-art WSSS methods.
\end{itemize}

In summary, the proposed CPC framework presents a unified strategy that combines language-derived semantic clustering with contrastive visual discrimination. Our work highlights the synergy between prompt-driven knowledge and contrastive representation learning, offering a new perspective on bridging vision-language priors and fine-grained segmentation in weakly supervised settings.

\section{Related Work}

\subsection{WSSS with Image-level Labels}
Weakly Supervised Semantic Segmentation (WSSS) that relies solely on image-level annotations often initiates its training by generating pseudo masks derived from Class Activation Maps (CAM)~\cite{wu2025image,wu2025generative,zhang2025timeraf,wu2025prompt}. Nevertheless, since CAMs typically capture only the most salient object parts, various efforts have emerged to overcome this limitation. To enrich the activated regions, techniques such as erasing~\cite{wei2017object,liu2024pcsformer}, attention accumulation during training~\cite{jiang2019integral}, and semantic consistency mining across images~\cite{sun2020mining} have been proposed. Additionally, some approaches~\cite{lee2021railroad,yao2021non} incorporate saliency guidance to suppress irrelevant background and reveal under-highlighted object parts. Other solutions~\cite{chen2022self,du2022weakly,cao2023gradient,zhao2023weight,xu2019polygon,luo2022learning} focus on contrastive objectives between pixel embeddings and class prototypes to encourage complete region coverage. Moreover, the work in~\cite{chang2020weakly} enhances segmentation by injecting auxiliary category knowledge and refining the learned features through mining label priors in the training set.  

Recent methods have also integrated Vision Transformers (ViT) into WSSS pipelines. For example, MCTformer~\cite{xu2022multi} and AFA~\cite{ru2022learning} leverage ViT’s self-attention to generate localization cues, where MCTformer uses PSA~\cite{ahn2018learning} and AFA adds affinity-aware refinement modules. In contrast, ViT-PCM~\cite{rossetti2022max} discards CAM entirely and estimates pixel-level labels via patch embeddings with max pooling, although it may suffer from patch-level noise. Unlike these paradigms, our CPC framework introduces additional clustering cues to guide the segmentation process, allowing more robust differentiation between semantically related classes and improving overall performance.

\subsection{Vision Transformers for WSSS}

Vision Transformer (ViT)~\cite{dosovitskiy2020image} has demonstrated strong performance across various vision-related benchmarks~\cite{xie2024weakly,xu2019polygon,luo2022learning,liuimproving,liu2024fedbcgd,xie2024accurate,dosovitskiy2020image}. This progress has encouraged its adoption in the field of WSSS, where ViT-based methods are gradually replacing traditional CNN-based CAM generation techniques~\cite{xu2022multi,ru2022learning}. For instance, MCTformer~\cite{xu2022multi} utilizes ViT’s self-attention to construct class localization maps and applies PSA~\cite{ahn2018learning} to generate pseudo masks. AFA~\cite{ru2022learning} extends this idea by incorporating a multi-head self-attention module alongside an affinity propagation unit to strengthen mask quality through contextual reasoning. Nevertheless, ViT's tendency towards over-smoothing continues to limit its ability to preserve fine details. In response, ViT-PCM~\cite{rossetti2022max} proposes to bypass CAM generation entirely by relying on patch embeddings and max pooling to estimate dense pixel predictions, though it risks introducing noise due to potential patch misclassification. Unlike these prior strategies, our method leverages additional cluster-level information to improve class discrimination, thereby enhancing the WSSS pipeline's ability to handle visually similar categories and overcoming several limitations inherent to ViT-based segmentation methods.

\subsection{Prompt-based Language Models}
Prompt-based strategies have become a powerful mechanism to adapt pre-trained language models (PLMs) to downstream tasks by embedding task-specific cues directly into the input. Initial approaches often relied on manually crafted prompts tailored to particular generation objectives~\cite{brown2020language,raffel2020exploring,zou2021controllable,chensnow,FANG2025127397,jin2025sscm}. However, such manually designed prompts tend to suffer from limited generalizability and poor adaptability to unseen tasks. To overcome these issues, subsequent research introduced automatic prompt construction techniques~\cite{shin2020autoprompt}, which eliminate the need for manual engineering. Moreover, other methods have focused on continuous prompt tuning~\cite{liu2023gpt,li2021prefix}, which replaces discrete token sequences with trainable embeddings, offering more flexible and effective adaptation across diverse tasks. Recently, the scope of prompt-based PLMs has expanded into the vision domain—for example,~\cite{zhang2023prompt} demonstrates the benefit of prompts for few-shot learning in visual recognition. Differing from these studies, our method employs PLMs to synthesize diverse textual prompts aimed at enriching image descriptions. To our knowledge, this is the first work that exploits PLM-generated prompts to derive semantic cluster information specifically designed to enhance performance in the WSSS setting.

\section{Methodology}
\label{sec:method}

In this section, we present the overall architecture and key components of our proposed Contrastive Prompt Clustering framework. We begin in Sec.~\ref{sec:Overview} with a high-level overview of the entire pipeline, including its core structures and interactions. Then, in Sec.~\ref{sec:prompt}, we introduce our prompt-driven category clustering method, which leverages the powerful semantic reasoning capability of Large Language Models (LLMs) to automatically group semantically similar categories. These clusters enable the model to better model inter-class relationships and reduce semantic confusion. In Sec.~\ref{sec:token}, we describe how the clustered category tokens are integrated into the ViT backbone to enhance semantic discrimination. 

\tbl{
Finally, in Sec.~\ref{sec:PatchContrast}, we introduce a class-aware patch contrastive learning objective that reinforces intra-class consistency by pulling together high-confidence patch embeddings from the same class while pushing away low-confidence ones. This component complements prompt-based clustering by aligning local visual features with semantic category coherence.
}

\begin{figure*}[t] 
\begin{center}
   \includegraphics[width=1.0\linewidth]{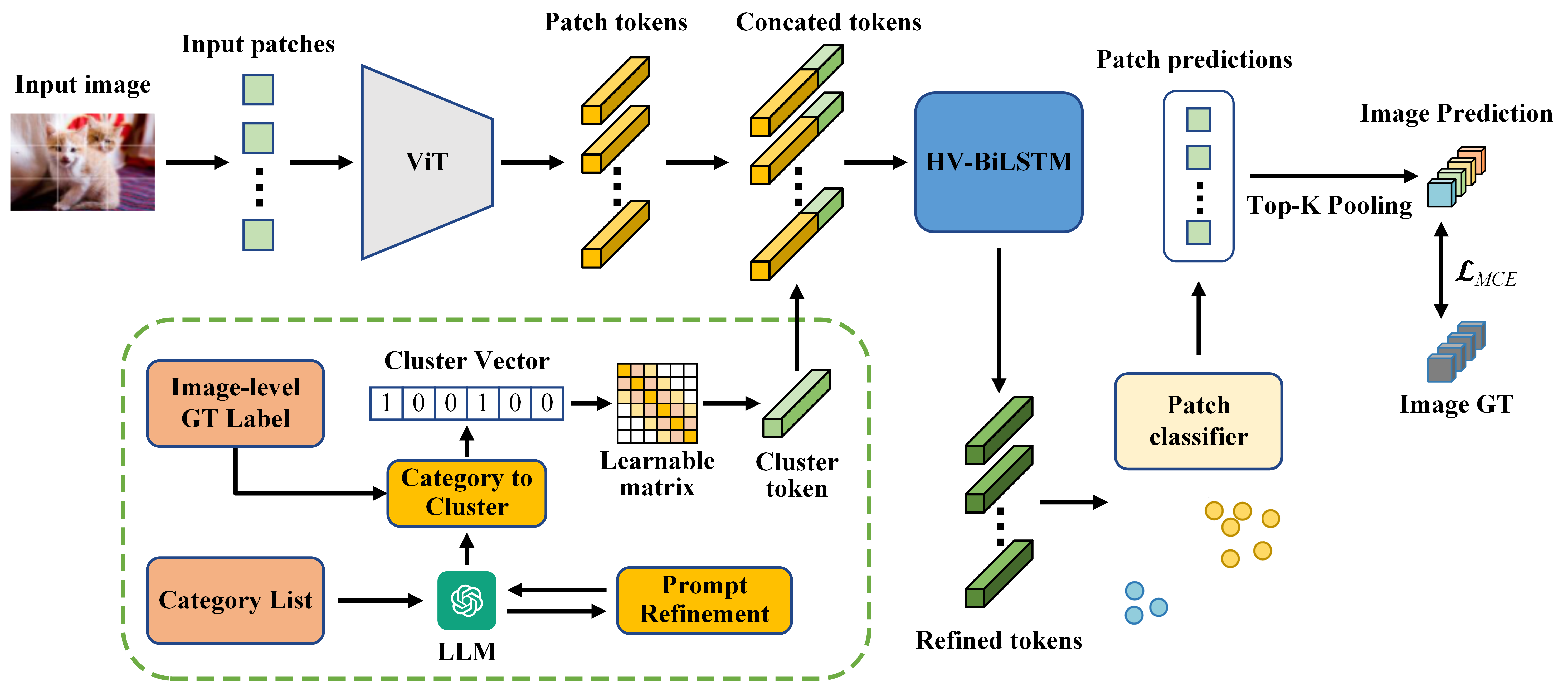}

   \caption{The overall framework of our proposed CPC is illustrated. First, a ViT backbone is used to extract patch-level tokens from the input image. Meanwhile, a predefined category list is fed into LLM to generate clusters of semantically related categories. Based on the image-level labels, a binary cluster vector is constructed, where 1 indicates membership in a cluster and 0 otherwise. This vector is multiplied with a learnable projection matrix to produce a cluster token that encodes semantic cluster information. The cluster token is then concatenated with the patch tokens and fed into a HV-BiLSTM module to produce refined patch embeddings. These embeddings are used in two branches: (1) a patch classifier outputs patch-level class probabilities, which are aggregated via top-k pooling to obtain image-level predictions for computing the MCE loss; (2) a patch contrastive learning module enhances intra-class consistency by pulling together confident patches of the same class and pushing them away from uncertain patches. The final pseudo-labels are generated via CRF post-processing and used to train a segmentation network.}
    \label{fig:CPC}
\end{center}
% \vspace{-0.3cm}
\end{figure*}

\subsection{Overall Framework} \label{sec:Overview}

As illustrated in Fig.~\ref{fig:CPC}, we provide a comprehensive overview of our proposed \textbf{CPC} framework. The network is designed to predict patch-level semantic labels by injecting both visual features and semantic cluster priors into the representation learning process. Specifically, the original ViT patch tokens are concatenated with a set of cluster tokens derived from LLM-guided prompt clustering, resulting in enriched patch representations. These tokens are further refined using a HV-BiLSTM module~\cite{rossetti2022max} to better model inter-patch dependencies.

A lightweight MLP-based classifier then predicts the patch-to-category associations, generating coarse pseudo-labels. These initial predictions are post-processed with a Conditional Random Field (CRF)~\cite{krahenbuhl2011efficient} to enforce spatial consistency and semantic alignment, resulting in high-quality pseudo-labels.

\tbl{
In parallel, we apply a patch contrastive loss on the intermediate feature space to enforce intra-class compactness and improve the robustness of patch-level representations (see Sec.~\ref{sec:PatchContrast}). This contrastive module operates on confident patch pairs, promoting more discriminative and stable embeddings.
}

Finally, the refined pseudo-labels are used to train a DeepLabv2~\cite{chen2018encoder} segmentation network in a fully supervised manner, which produces the final dense semantic predictions. This two-stage training pipeline enables our framework to effectively leverage both language-guided clustering and feature-level supervision, achieving superior performance in weakly supervised semantic segmentation.

\begin{algorithm}[thbp]
\caption{{\small{Self-Refine category clusters generation}}}\label{alg:SR}
\SetKwInOut{Input}{Input}
\SetKwInOut{Output}{Output}

\Input{category list $l$, LLM $\mathcal{M}$, initial prompt $\pgen$, refine prompt $\refine$, stop condition $s()$, number of queries $R$}
\Output{ category clusters $z_t$}
\BlankLine
\tbl{// Step 1: Multi-query initialization with voting}\\
\tbl{Initialize result set $\mathcal{Z} = \{\}$}\\
\tbl{\For{$r = 1$ \KwTo $R$}{
    $z^{(r)} = \mathcal{M}(\pgen \parallel l)$ with $T{=}0$\;
    $\mathcal{Z} = \mathcal{Z} \cup \{z^{(r)}\}$\;
}}
\tbl{$z_0 = \text{most frequent element in } \mathcal{Z}$}\;
\BlankLine

// Step 2: Iterative refinement
\For { iteration $t \in \{0, 1, \ldots\}$}{
    $z_{t+1} = \mathcal{M}(\refine \parallel z_t)$\;
    \If{$s(z_{t+1}, z_{t}, \ldots, z_{0})$}{
        \textbf{break}\;
    }
}
\Return $z_t$\;
\end{algorithm}

\subsection{Self-Refine Prompt for Cluster} \label{sec:prompt}

The motivation behind clustering categories arises from our observation that, in previous works, models are typically designed to maximize the distance between different classes while minimizing the distance within the same class. This approach often leads to the assumption that the distances between all classes are equidistant, ignoring their intrinsic semantic similarities. For instance, a model might perceive the distance from ``cat" to ``dog" as equal to the distance from ``cat" to ``car," despite the fact that ``cat" and ``dog" share significantly more similarities as animals, whereas "cat" and "car" are entirely distinct objects. To address this limitation and better capture the nuanced relationships between classes, we propose to abstract more granular cluster information at the class level. Specifically, we harness the powerful capabilities of LLMs to assign distinct cluster tags to these classes. However, due to the inherent instability of LLMs~\cite{madaan2024self}, the generated cluster tags may lack sufficient reliability. 

\tbl{To mitigate this issue and enhance robustness, we query the LLM $R$ times using a fixed prompt schema and temperature $T{=}0$, generating $R$ candidate cluster partitions for the category list. We then perform frequency-based voting across these partitions and select the most commonly occurring cluster configuration as the final result. The number of clusters is determined by the majority count observed across all queries, enabling a data-driven estimation of cluster granularity.} 

We integrate this self-refining prompt method into our clustering pipeline, as elaborated in Algo.~\ref{alg:SR} and visually represented by the Category Cluster module in Fig.~\ref{fig:CPC}. 
\tbl{Here, $R$ denotes the number of repeated LLM queries for initialization, $\mathcal{Z}$ represents the set of all sampled cluster results $z^{(r)}$, and the function $\texttt{most\_frequent}(\cdot)$ selects the most commonly occurring configuration among them. These partitions are generated under deterministic decoding ($T{=}0$) and aggregated via majority voting to identify the most stable clustering outcome.}

\begin{figure*}[t]
\centering
\includegraphics[width=0.8\linewidth]{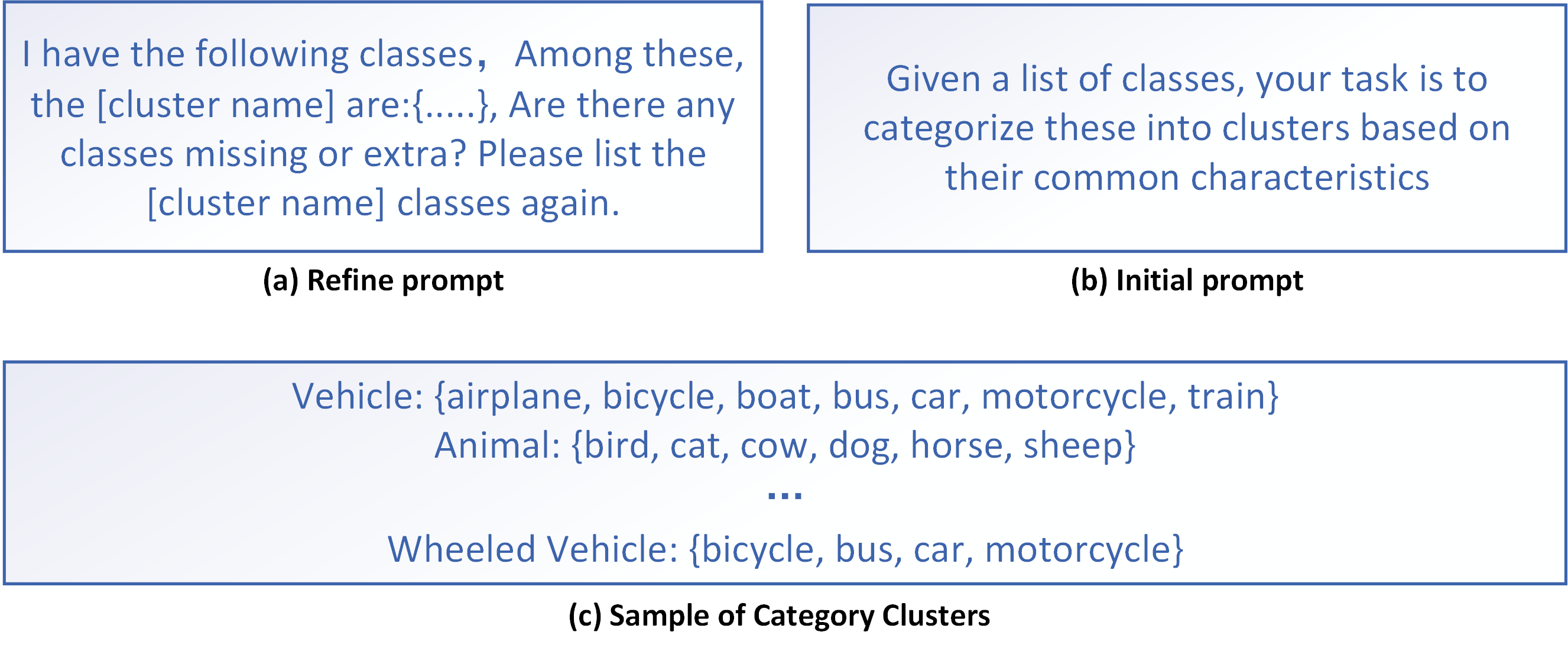}
\caption{The proposed Prompt strategy.(a) is the refined prompt template; (b) is the initial prompt template; (c) is the samples of category clusters.}
\label{fig:prompt}
\vspace{-0.3cm} 
\end{figure*}

\tbl{The clustering generation process is thus transformed into a consensus-driven mechanism, where the initial phase collects candidate partitions via multiple deterministic LLM queries.} Starting with a comprehensive category list $l$ extracted from the dataset, we initially utilize a clustering prompt $\pgen$ (depicted in Fig.~\ref{fig:prompt}(b)) in conjunction with GPT-4o ($\mathcal{M}$) to generate preliminary category clusters. This phase lays the groundwork by establishing baseline associations between semantic concepts.

Following this, the algorithm enters an optimization loop, where a refinement prompt $\refine$ (illustrated in Fig.~\ref{fig:prompt}(a)) facilitates iterative enhancements. At each iteration $t$, the language model $\mathcal{M}$ processes the current cluster configuration $z_t$ through the refinement template to produce an improved version $z_{t+1}$. The stopping condition $s()$ evaluates convergence by monitoring cluster stability across three consecutive iterations (i.e., when $z_{t+1} = z_t = z_{t-1}$), ensuring that the algorithm terminates only when the semantic groupings reach a stable equilibrium. This mechanism guarantees that the final clusters are both semantically coherent and robust.

Through this iterative generation-refinement process, the method systematically consolidates semantically coherent clusters, progressively enhancing their accuracy and consistency. Fig.~\ref{fig:prompt}(c) illustrates final cluster samples that exhibit significantly improved categorical cohesion compared to the initial groupings, highlighting the effectiveness of the refinement mechanism. The dynamic nature of this process enables the system to refine semantic relationships through successive reasoning steps, ultimately yielding more precise and meaningful cluster configurations.

 \subsection{Category Cluster Tokens for WSSS} \label{sec:token}

To enhance the richness of category information, we integrate cluster tokens as supplementary components within the training framework. These tokens enable the network to more effectively capture and utilize similarity relationships across diverse categories. Specifically, we concatenate ViT tokens with cluster tokens to construct the final input tokens, thereby optimizing the model's capacity to leverage inter-class similarities. For example, while the label ``car" represents a distinct category, the label ``cat" exhibits greater similarity to "dog" due to shared semantic attributes. Consequently, both ``cat" and ``dog" are grouped into clusters such as ``animal" or ``pet," reflecting their common contextual characteristics and enabling the model to better understand and exploit these relationships during training.

The pipeline of our proposed method is depicted in Fig.~\ref{fig:CPC}. The input image $X$, with dimensions $\mathbb{R}^{h \times w \times 3}$, is fed into a Vision Transformer (ViT) encoder to extract patch tokens $F_{v} \in \mathbb{R}^{s \times e}$, where $s = (n/d)^2$ denotes the number of tokens and $e$ represents the dimensionality of each token. In practice, all input images are resized to consistent dimensions $h$ and $w$, such that $n = h = w$, and $d$ corresponds to the patch size. To incorporate cluster-level information, we design a cluster vector $u \in \mathbb{R}^L$ to indicate the cluster affiliations of $X$. Each element $u_i \in \{0, 1\}$ signifies the presence (1) or absence (0) of cluster category $i$, where $L$ represents the total number of category clusters. This cluster vector $u$ is then mapped to a cluster token $F_c \in \mathbb{R}^{H}$ through a learnable matrix $G \in \mathbb{R}^{L \times H}$, computed as follows:

\begin{equation}
\begin{aligned}
    F_c = u^T \cdot G.
\end{aligned}
\end{equation}

Subsequently, we concatenate the cluster token $F_c$ with each patch token, resulting in the combined tokens $F_{in} \in \mathbb{R}^{s \times (e+H)}$. This integration enables the WSSS framework to more effectively perceive and utilize cluster-level information, thereby significantly enhancing the overall segmentation performance by leveraging the enriched contextual relationships between categories.

We employ a HV-BiLSTM module to refine the concatenated tokens $F_{in}$, producing refined tokens $F_{out} \in \mathbb{R}^{s \times (e+H)}$ with identical dimensions. Following this refinement, we introduce a one-layer Multi-Layer Perceptron (MLP) patch classifier, parameterized by a weight matrix $W \in \mathbb{R}^{(e+H) \times C}$, where $C$ denotes the total number of classes. By applying the MLP and SoftMax operations, we generate predictions $Z \in \mathbb{R}^{s \times C}$ from the patch classifier, formulated as follows:

\begin{equation}
\begin{aligned}
    Z=\text{softmax}(F_{out}W),
\end{aligned}
\end{equation}

the variable $Z$ represents the patch-level predictions for semantic segmentation. To align these patch predictions with the image-level supervision required for training, it is necessary to transform them into image class predictions. This conversion is essential for accurate loss computation and ensuring that the predictions are consistent with the provided image-level labels. To achieve this, we employ a Top-K pooling mechanism to aggregate the patch predictions and derive the image class predictions $p_c$ for class $c$, as follows:

\begin{equation} 
\begin{aligned}
    p_c = \frac{1}{k} \sum_{i=1}^{k} \text{Top-k}(Z_{j}^{c}) ~~and ~~j \in \{1, \ldots, s\},
\end{aligned}
\end{equation}

where $\text{Top-k}(\cdot)$ denotes the operation of selecting the top $k$ patches with the highest prediction values for class $c$. This mechanism ensures that the final image-level predictions are robust and not disproportionately influenced by any anomalous or outlier patches.

\subsection{Patch Contrastive Learning for Intra-Class Consistency} \label{sec:PatchContrast}

While the proposed Prompt-guided Clustering module effectively captures inter-class semantic relationships, it remains crucial to enhance the consistency of patch embeddings within each class to ensure robust predictions. To address this, we introduce a Patch Contrastive Error (PCE) loss, which explicitly pulls together high-confidence patch embeddings of the same class while simultaneously pushing them away from low-confidence embeddings within that class.

Let $F_{out}$ denote the refined patch embeddings and $Z$ denote the patch prediction scores from the MLP classifier. For a given category $c$, we define high-confidence patches as $\mathcal{P}^{c}_{high} = \{F^{i}_{out} \mid Z_i^c > \epsilon\}$ and low-confidence patches as $\mathcal{P}^{c}_{low} = \{F^{i}_{out} \mid Z_i^c < 1 - \epsilon\}$, where $\epsilon$ is a predefined confidence threshold.

We measure similarity between two embeddings using cosine similarity:
\begin{equation}
S(F_{out}^{i}, F_{out}^{j}) = \frac{F_{out}^{i} \cdot F_{out}^{j}}{\|F_{out}^{i}\|\|F_{out}^{j}\|},
\end{equation}
which is normalized to $[0,1]$ by:
\begin{equation}
\bar{S}(F_{out}^{i}, F_{out}^{j}) = \frac{1 + S(F_{out}^{i}, F_{out}^{j})}{2}.
\end{equation}

The overall Patch Contrastive Error for category $c$ is then formulated as:
\begin{equation}
\begin{aligned}
\mathcal{L}_{PCE}^{c} &= \frac{1}{N_{pair}^{+}} \sum_{i=1}^{|\mathcal{P}^{c}_{high}|} \sum_{j=1, j \neq i}^{|\mathcal{P}^{c}_{high}|} (1 - \bar{S}(F_{high}^{i}, F_{high}^{j})) \\
&+ \frac{1}{N_{pair}^{-}} \sum_{m=1}^{|\mathcal{P}^{c}_{high}|} \sum_{n=1}^{|\mathcal{P}^{c}_{low}|} \bar{S}(F_{high}^{m}, F_{low}^{n}),
\end{aligned}
\end{equation}
where $N_{pair}^{+}$ and $N_{pair}^{-}$ denote the number of high-high and high-low patch pairs, respectively. The first term minimizes intra-class embedding variance, while the second term penalizes similarity between confident and uncertain regions, effectively pushing them apart.

By incorporating this class-aware contrastive learning module, the model benefits from both global semantic alignment via CPC and local feature consistency within each class, resulting in more accurate and coherent segmentation outcomes.
\subsection{Overall Loss} \label{sec:loss}

We minimize the multi-label classification prediction error (MCE) between the predicted image-level labels $p_c$ and the ground truth image labels $y_c$, ensuring that the model's predictions align closely with the actual annotations. This optimization process is formalized as follows:
\begin{equation} 
\begin{aligned}\label{eq:MCE}
\mathcal{L}_{MCE}&=\frac{1}{C}\sum_{c\in \mathcal{C}}{BCE(y_c,p_c)}\\
&=-\frac{1}{C}\sum_{c\in \mathcal{C}}{y_c\log(p_c)+(1-y_c)\log(1-p_c)},
\end{aligned}
\end{equation}
where $C$ represents the number of classes within the dataset and BCE denotes the binary cross-entropy loss.

\tbl{
To further enhance the intra-class consistency of patch embeddings, we introduce an additional class-aware contrastive loss, $\mathcal{L}_{PCE}$, as described in Sec.~\ref{sec:PatchContrast}. This loss pulls together high-confidence patches from the same class while pushing them away from low-confidence patches. It complements $\mathcal{L}_{MCE}$ by improving patch-level discrimination and spatial consistency, which are critical under weak supervision. The overall loss function is defined as:
\begin{equation}
\mathcal{L}_{total} = \mathcal{L}_{MCE} + \lambda_{PCE} \sum_{c \in \mathcal{C}} \mathcal{L}_{PCE}^{c},
\end{equation}
where $\lambda_{PCE}$ is a weighting hyperparameter controlling the influence of the contrastive term.
}

During the inference stage, we utilize the trained patch classifier to generate patch-level softmax class predictions $Z$. Following this, an interpolation algorithm is applied to $Z$ to upsample and obtain pixel-level softmax predictions for the entire input image. Finally, an argmax operation is performed on the interpolated $Z$ along the class dimension $C$ to assign the class pseudo-label to each pixel, resulting in the final semantic segmentation output.

To elaborate further, during training our model focuses on reducing the discrepancy between the predicted probabilities and the true labels across all classes. The loss $\mathcal{L}_{MCE}$ is computed by averaging the binary cross-entropy loss over the set of classes $\mathcal{C}$, ensuring that every class contributes equally regardless of its frequency in the dataset. This uniform treatment helps address any potential class imbalance and encourages the network to learn robust representations for both frequent and rare classes.

\tbl{
Meanwhile, the contrastive loss $\mathcal{L}_{PCE}$ enforces compactness among high-confidence patch embeddings of the same category, acting as a structural regularizer that complements the global guidance from category clusters.
}

At inference time, the process shifts focus from training to generating meaningful predictions for new images. Initially, the trained patch classifier produces softmax outputs at the patch level, resulting in a prediction map $Z$. Since these predictions are at a lower resolution corresponding to the patches, an interpolation algorithm is applied to upscale $Z$ to the original image resolution. This step ensures that the spatial details captured by the patch classifier are retained and properly represented in the final output. Finally, to convert these continuous softmax predictions into discrete class labels, an argmax operation is performed along the class dimension $C$. Each pixel is then assigned the label corresponding to the highest probability, yielding the final pseudo-label map for semantic segmentation. This complete pipeline enables our model to achieve precise segmentation results.

\section{Experiments}
\label{sec:Experiments}

In this section, we present our experimental settings and evaluation protocols. We compare our proposed method, Contrastive Prompt Clustering (CPC), with state-of-the-art weakly supervised semantic segmentation (WSSS) approaches on the PASCAL VOC 2012~\cite{everingham2010pascal} and MS COCO 2014~\cite{lin2014microsoft} benchmarks. Finally, we perform ablation studies to validate the contributions of each component in our framework.

\subsection{Experimental Settings}

\textbf{Datasets and Evaluation Metric.}  
We evaluate the effectiveness of the proposed framework on two widely adopted benchmarks for weakly supervised semantic segmentation: PASCAL VOC 2012~\cite{everingham2010pascal} and MS COCO 2014~\cite{lin2014microsoft}. The PASCAL VOC 2012 dataset comprises 20 foreground object categories and one background class. Following standard practice, we augment the original training set with additional annotations from the Semantic Boundaries Dataset (SBD)~\cite{hariharan2011semantic}, resulting in 10,582 training images annotated with image-level labels and 1,449 images reserved for validation. MS COCO 2014 provides a more challenging evaluation setting, containing 81 object categories with significant intra-class variation and complex scene compositions. In accordance with prior WSSS literature, we employ approximately 82,000 images with image-level supervision for training and 40,000 images from the validation set for performance evaluation.

For quantitative assessment, we adopt the standard mean Intersection over Union (mIoU) metric, which measures the average IoU across all semantic classes in the validation set. mIoU offers a balanced evaluation of both region-wise prediction accuracy and boundary localization, making it particularly suitable for benchmarking semantic segmentation performance under weak supervision.

\textbf{Implementation Details.}  
In our experiments, we utilize the ViT-B/16 model as the encoder, owing to its demonstrated ability to capture rich visual representations. Each image is resized to $384 \times 384$ during training~\cite{kolesnikov2016seed} to ensure consistent input resolution and is subsequently divided into $24 \times 24$ non-overlapping patches of size $16 \times 16$. This patch-wise representation enables the transformer architecture to model both local and global contextual information through self-attention mechanisms.

The model is trained for 50 epochs with a batch size of 16 on two NVIDIA RTX 4090 GPUs. We adopt the Adam optimizer with an initial learning rate of $10^{-3}$ for the first two epochs, followed by a reduced learning rate of $10^{-4}$ for the remaining training duration. This two-stage schedule facilitates both rapid convergence and stable optimization.

\begin{figure*}[t]
\centering
\includegraphics[width=0.9\linewidth]{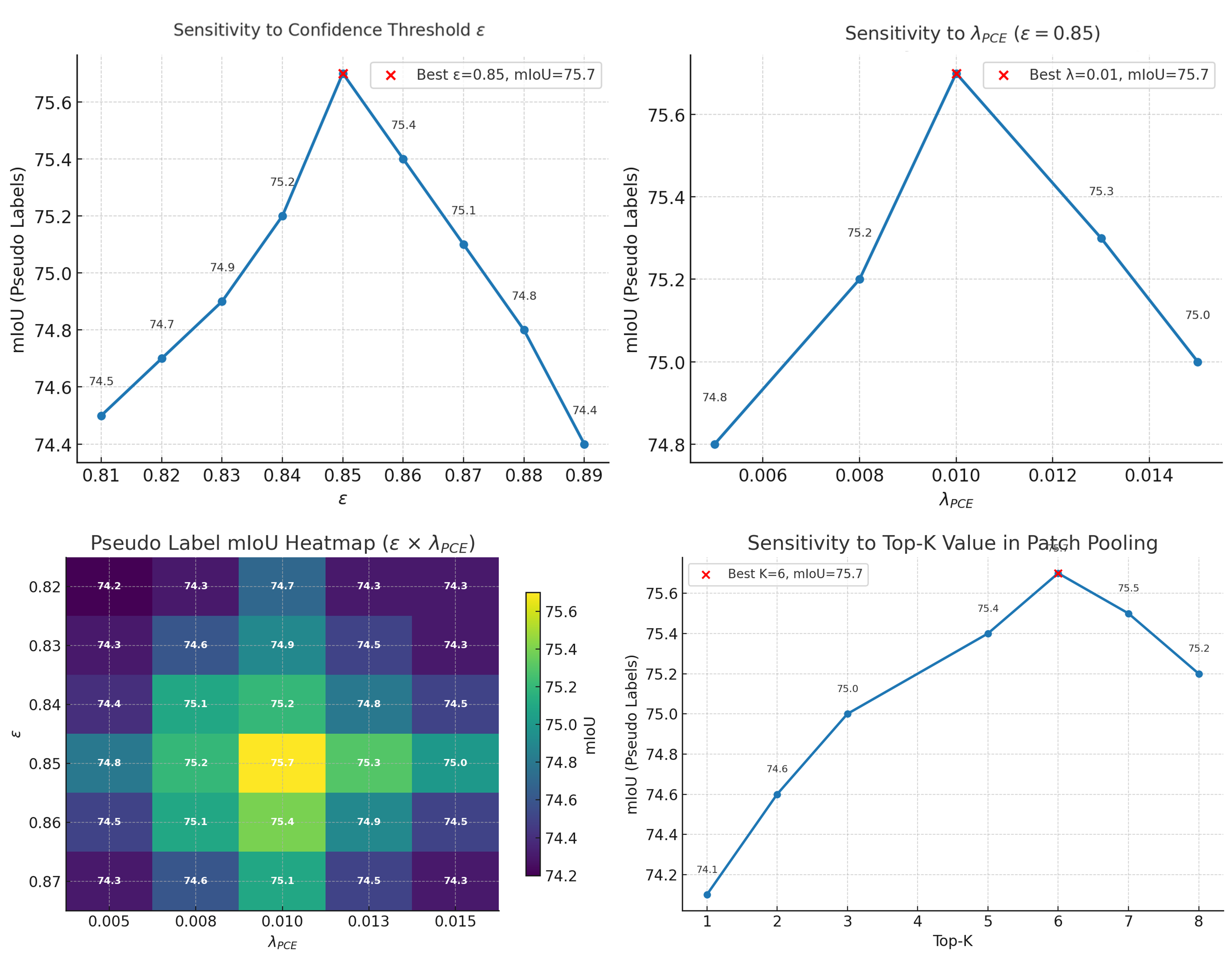}
\vspace{-0.1cm}
\caption{Qualitative visualization of segmentation results. Our CPC framework demonstrates clearer object boundaries and more accurate region predictions compared to baselines.}
\label{fig:param}
\vspace{-0.1cm}
\end{figure*}

During training, Top-K pooling with $k = 6$ is applied to the patch classifier outputs to aggregate the most salient patch-level activations, yielding image-level predictions supervised by the multi-label classification loss and query count of $R=10$ is used to ensure stability in the LLM results.
\tbl{Additionally, we incorporate a patch contrastive learning module to improve intra-class feature compactness, where the contrastive loss is computed between high- and low-confidence patch embeddings for each class. Specifically, the high-confidence and low-confidence patches are selected using a predefined threshold $\epsilon = 0.85$.
The final training objective combines the classification loss and the contrastive loss, with the latter weighted by a hyperparameter $\lambda_{PCE} = 0.01$.}

For inference, input images are upsampled to $960 \times 960$ to generate high-resolution predictions. The patch classifier produces class probability maps, which are subsequently refined using Conditional Random Fields (CRF)~\cite{krahenbuhl2011efficient} to improve spatial coherence and boundary accuracy. These refined pseudo-labels are employed to supervise a DeepLab V2~\cite{chen2018encoder} segmentation model, which leverages atrous convolutions to capture multi-scale contextual dependencies, thereby producing the final dense semantic segmentation output.

\begin{table}[ht]
\centering
\caption{Pseudo Label Performance Comparison (mIoU) on PASCAL VOC 2012.}
\label{tab:vocbpm}
\begin{adjustbox}{width=0.9\linewidth}
\begin{tabular}{@{}lllll@{}}
\toprule
method & Pub. & Backbone & mIoU \\
\midrule
AdvCAM~\cite{lee2022anti} & PAMI22 & V2-RN101 & 69.9 \\
SIPE~\cite{chen2022self} & CVPR22 & ResNet50 & 58.6 \\
AFA~\cite{ru2022learning} & CVPR22 & MiT-B1 & 66.0 \\
ViT-PCM~\cite{dosovitskiy2020image} & ECCV22 & ViT-B/16 & 71.4 \\
ToCo~\cite{ru2023token} & CVPR23 & ViT-B/16 & 72.2 \\
USAGE~\cite{Peng_2023_ICCV} & ICCV23 & ResNet38 & 72.8 \\
FPR~\cite{chen2023fpr} & ICCV23 & ResNet38 & 68.5 \\
ToCo~\cite{ru2023token} & CVPR23 & ViT-B/16 & 70.5 \\
SFC~\cite{zhao2024sfc} & AAAI24 & ViT-B/16 & 73.7 \\
APC~\cite{wu2025adaptive} & EAAI25& ViT-B/16 & 74.6 \\
\textbf{CPC} & \textbf{Ours} & ViT-B/16 & \textbf{75.7} \\
\bottomrule
\end{tabular}
\end{adjustbox}
\end{table}

\subsection{Comparison with State-of-the-arts}
\textbf{Pseudo Label Performance Comparison .}  
The proposed CPC framework effectively exploits the latent semantic correlations among visually similar categories, enabling the model to better disambiguate closely related classes under weak supervision. By introducing prompt-guided category clusters, the model captures shared feature structures that are often overlooked in conventional approaches. This facilitates the learning of more discriminative and semantically aware patch representations, ultimately leading to improved pseudo-label quality.

In addition to the inter-class clustering, our framework integrates a patch-level contrastive loss to further enhance intra-class consistency, which contributes to more coherent and robust pixel-level predictions. Together, these components allow the model to more accurately distinguish between ambiguous regions and reduce misclassification errors caused by overlapping semantics. As shown in Tab.~\ref{tab:vocbpm}, our method achieves superior mIoU performance compared to a range of recent state-of-the-art methods, including both convolution-based and transformer-based backbones. Notably, CPC outperforms strong baselines such as ToCo~\cite{ru2023token} and SFC~\cite{zhao2024sfc}, demonstrating the effectiveness of our dual-level learning strategy in producing high-quality pseudo labels.

\begin{table}[ht] 
\centering
\caption{Semantic Segmentation Performance Comparison (mIoU) on PASCAL VOC 2012.}
\label{tab:vocseg}
\begin{adjustbox}{width=0.9\linewidth}
\begin{tabular}{@{}lllll@{}}
\toprule
Model & Pub. & Backbone & Val \\
\midrule
USAGE~\cite{Peng_2023_ICCV} & ICCV23 & ResNet38 & 71.9 \\
SAS~\cite{kim2023semantic} & AAAI23 & ViT-B/16 & 69.5 \\
MCTformer~\cite{xu2022multi} & CVPR22 & DeiT-S & 61.7 \\
SIPE~\cite{chen2022self} & CVPR22 & ResNet50 & 58.6 \\
ViT-PCM~\cite{dosovitskiy2020image} & ECCV22 & ViT-B/16 & 69.3 \\
AFA~\cite{ru2022learning} & CVPR22 & MiT-B1 & 63.8 \\
ToCo~\cite{ru2023token} & CVPR23 & ViT-B/16 & 70.5 \\
TSCD~\cite{Xu_Wang_Sun_Xu_Meng_Zhang_2023} & AAAI23 & MiT-B1 & 67.3 \\
FPR~\cite{chen2023fpr} & ICCV23 & ResNet38 & 70.0 \\
SFC~\cite{zhao2024sfc} & AAAI24 & ViT-B/16 & 71.2 \\
IACD~\cite{wu2024image} & ICASSP24 & ViT-B/16 & 71.4 \\
APC~\cite{wu2025adaptive} & EAAI25& ViT-B/16 & 72.3 \\
\textbf{CPC} & \textbf{Ours} & ViT-B/16 & \textbf{73.4} \\
\bottomrule
\end{tabular}
\end{adjustbox}
\end{table}

\begin{figure*}[t]
\centering
\includegraphics[width=0.9\linewidth]{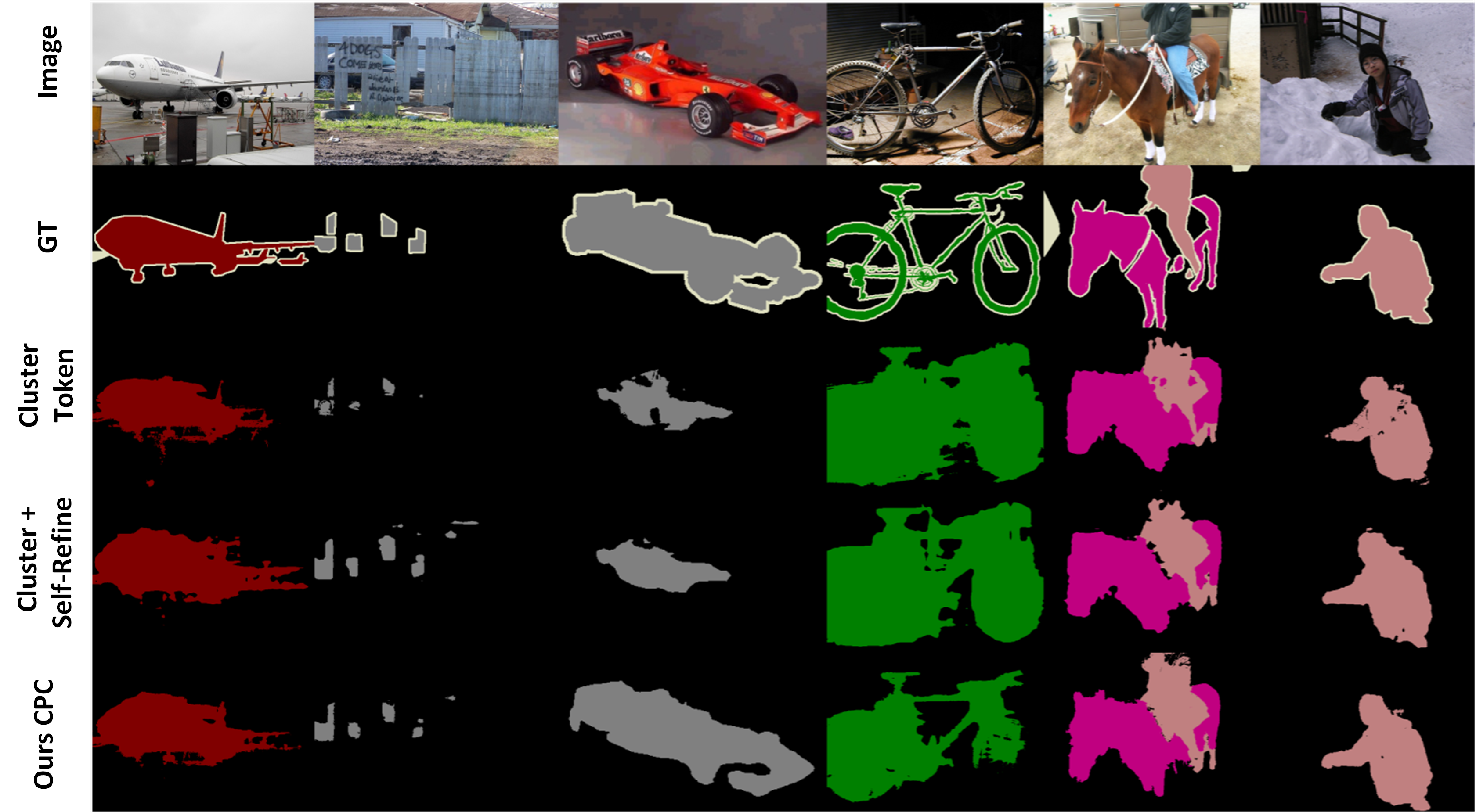}
\vspace{-0.1cm}
\caption{Qualitative visualization of segmentation results. Our CPC framework demonstrates clearer object boundaries and more accurate region predictions compared to baselines.}
\label{fig:result}
\vspace{-0.1cm}
\end{figure*}

\begin{table}[t] 
\centering
\caption{Semantic Segmentation Performance Comparison (mIoU) on MS COCO 2014.}
\label{tab:cocoseg}
\begin{adjustbox}{width=0.9\linewidth}
\begin{tabular}{@{}llll@{}}
\toprule
Model & Pub. & Backbone & mIoU (\%) \\
\midrule
MCTformer~\cite{xu2022multi} & CVPR22 & Resnet38 & 42.0 \\
ViT-PCM~\cite{dosovitskiy2020image} & ECCV22 & ViT-B/16 & 45.0 \\
SIPE~\cite{chen2022self} & CVPR22 & Resnet38 & 43.6\\
TSCD~\cite{Xu_Wang_Sun_Xu_Meng_Zhang_2023} & AAAI23 & MiT-B1 & 40.1 \\
SAS~\cite{kim2023semantic} & AAAI23 & ViT-B/16 & 44.5 \\
FPR~\cite{chen2023fpr} & ICCV23 & ResNet38 & 43.9 \\
ToCo~\cite{ru2023token} & CVPR23 & ViT-B & 42.3 \\
SFC~\cite{zhao2024sfc} & AAAI24 & ViT-B/16 & 44.6 \\
IACD~\cite{wu2024image} & ICASSP24 & ViT-B/16 & 44.3 \\
PGSD~\cite{hao2024prompt} & TCSVT24 & ViT-B/16 & 43.9 \\
APC~\cite{wu2025adaptive} & EAAI25& ViT-B/16 & 45.7 \\
\textbf{CPC} & \textbf{Ours} & ViT-B/16 & \textbf{46.7} \\
\bottomrule
\end{tabular}
\end{adjustbox}
\end{table}

\textbf{Semantic Segmentation Results.}  
To evaluate the effectiveness of our proposed pseudo-label generation strategy, we adopt DeepLab V2~\cite{chen2018encoder} as the segmentation backbone and train it using the pseudo labels generated by our CPC framework. We report semantic segmentation performance on the PASCAL VOC 2012 validation set and compare it with recent state-of-the-art WSSS approaches, as presented in Tab.~\ref{tab:vocseg}. Our CPC framework yields superior segmentation performance, achieving an mIoU of 72.2\% and outperforming prior transformer-based methods such as ToCo~\cite{ru2023token}, SFC~\cite{zhao2024sfc}, and IACD~\cite{wu2024image}. These results confirm the advantage of combining inter-class category clustering with intra-class contrastive learning for generating high-quality pseudo labels. The enhanced quality of pseudo labels leads to clearer delineation of complex object structures and reduced confusion between visually similar classes.

\textbf{Visualization of Result.} 
To further assess the effectiveness of our method, we present qualitative segmentation results in Fig.~\ref{fig:result}. Compared to existing state-of-the-art approaches, our CPC framework produces significantly more accurate and coherent segmentation outputs, particularly in regions characterized by fine-grained object boundaries and semantic overlaps. As illustrated, CPC effectively mitigates common segmentation errors such as over-smoothing and misclassification, resulting in clearer object contours and improved region integrity. These observations align well with the quantitative improvements and confirm the model's enhanced ability to capture precise semantic structures.

\begin{figure}[t]
\centering
\includegraphics[width=1.0\linewidth]{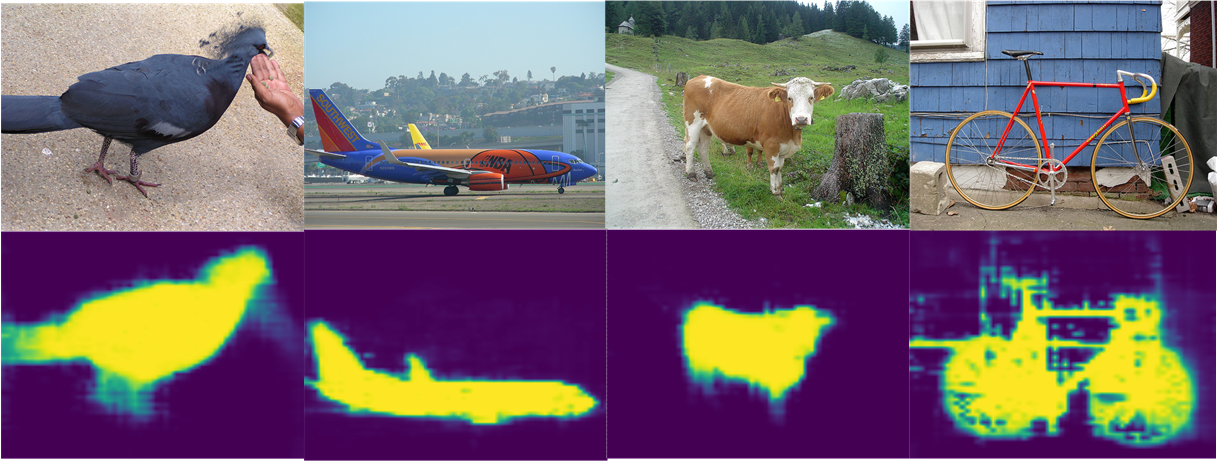}
\caption{Visualization of class-specific activation heatmaps. Yellow pixels indicate higher predicted probability for the target class, highlighting the model's spatial confidence.}
\label{fig:heatmap}
\end{figure}

In addition, we provide heatmap visualizations in Fig.~\ref{fig:heatmap} to interpret the model’s confidence across spatial locations. In these maps, pixel intensity reflects the predicted probability of belonging to the target class, with yellow indicating higher confidence. The results show that our model exhibits strong activations in object centers and smooth gradients toward object boundaries, indicating a well-calibrated spatial awareness. Notably, CPC yields sharper and more localized class-specific activations compared to prior methods, highlighting its superior ability to preserve semantic focus and reduce ambiguity at transition regions. These findings further validate that our approach enhances both object localization and semantic consistency across diverse visual scenes.

\subsection{Ablation Studies}

\begin{table}[ht]
\centering
\caption{Ablation studies on the impact of Self Refine, cluster token, and contrastive learning on PASCAL VOC 2012 Val}
\label{tab:ablation}
\begin{adjustbox}{width=1.0\linewidth}
\begin{tabular}{cccccc}
\toprule
Original Framework &  Cluster Token & Self Refine & Contrastive Learning & mIoU \\
\midrule
\checkmark &  &  &  & 68.6\% \\
\checkmark & \checkmark &  &  & 71.1\% \\
\checkmark & \checkmark & \checkmark &  & 72.2\% \\
\checkmark & \checkmark & \checkmark & \checkmark & \textbf{73.4\%} \\
\bottomrule
\end{tabular}
\end{adjustbox}
\end{table}
To validate the contribution of each core component in our framework, we conduct ablation studies on the PASCAL VOC 2012 validation set. \tbl{As summarized in Tab.~\ref{tab:ablation}, we evaluate the individual and combined effects of Self Refine, cluster token, and contrastive learning. Starting from a baseline ViT that uses only patch tokens, we first add the cluster token derived from GPT-generated category groupings. This token captures high-level semantic relations across categories and improves the model’s ability to reason about inter-class similarity, resulting in a +2.5\% mIoU gain. Next, we introduce the Self Refine mechanism, which performs multiple LLM queries followed by refinement, providing a more stable and semantically consistent cluster representation. This contributes an additional +1.1\% improvement. Finally, we incorporate a patch-level contrastive learning module that enforces intra-class consistency and encourages better separation of uncertain regions, leading to the highest mIoU of 73.4\%.}

\tbl{To evaluate the effect of query count $R$ in our Self-Refine clustering module, we conduct an ablation study focused on how different values of $R$ influence the stability of generated category clusters and the final segmentation performance. For each small $R$ (e.g., 2 or 6), we run the entire pipeline three times with independently sampled clusters and report the average mIoU and standard deviation. For large $R$ (e.g., 10), we observe consistent cluster outputs after majority voting and report the mIoU from a single run. As shown in Tab.~\ref{tab:ablation_r}, increasing $R$ improves cluster stability and leads to better and more reliable segmentation results.}

\begin{table}[ht]
\centering
\caption{Impact of the number of LLM queries $R$ on cluster stability and Semantic Segmentation Performance on PASCAL VOC 2012}
\label{tab:ablation_r}
\begin{adjustbox}{width=0.85\linewidth}
\begin{tabular}{cccc}
\toprule
\textbf{$R$} & \textbf{Cluster Stability} & \textbf{mIoU (avg ± std)} & \textbf{Note} \\
\midrule
2  & \ding{55} Unstable             & 73.0 ± 0.2 & Avg of 3 runs \\
6  & \ding{55} Occasionally Stable  & 73.1 ± 0.1 & Avg of 3 runs \\
10 & \ding{51} Stable               & \textbf{73.4} & Voted cluster, 1 run \\
\bottomrule
\end{tabular}
\end{adjustbox}
\end{table}

\tbl{These results demonstrate that a sufficiently large query count $R$ is crucial for ensuring semantically stable cluster assignments and reproducible segmentation performance. While smaller $R$ values can sometimes yield reasonable results, their non-deterministic nature introduces performance fluctuations and undermines robustness. Our final model adopts $R=10$ to balance cluster quality and computational efficiency.}

\section{Conclusion}
We propose CPC, a novel framework for WSSS. Unlike prior methods that process each category independently, CPC leverages LLM-generated category clusters to model inter-class semantic relationships. These clusters are encoded as learnable tokens and injected into the ViT backbone to enhance the model's ability to distinguish semantically similar classes. In addition, we introduce a class-aware patch contrastive loss to improve intra-class feature consistency. Combined, these components significantly improve pseudo-label quality and achieve state-of-the-art performance using only image-level supervision.

Beyond individual module effectiveness, our framework highlights the synergy between high-level semantic reasoning from language models and fine-grained visual discrimination through contrastive learning. This hierarchical design enables the model to simultaneously leverage global category relationships and local feature consistency, addressing long-standing challenges in WSSS such as semantic ambiguity and over-smoothing. Extensive experiments on PASCAL VOC and MS COCO demonstrate the robustness and generality of our method across diverse datasets. We believe CPC provides a promising foundation for future research at the intersection of vision and language in weakly supervised learning scenarios.

\section{Acknowledgements}
This work was supported by the National Natural Science Foundation of China (No. 62471405, 62331003, 62301451), Suzhou Basic Research Program (SYG202316) and XJTLU REF-22-01-010, XJTLU AI University Research Centre, Jiangsu Province Engineering Research Centre of Data Science and Cognitive Computation at XJTLU and SIP AI innovation platform (YZCXPT2022103).
\bibliography{mybib}

\end{document}